\pdfoutput=1

\documentclass[11pt]{article}

\usepackage[preprint]{acl}

\usepackage{times}
\usepackage{latexsym}

\usepackage[T1]{fontenc}

\usepackage[utf8]{inputenc}

\usepackage{microtype}

\usepackage{inconsolata}

\usepackage{amsmath}

\usepackage{graphicx}

\usepackage{multirow} 

\usepackage{booktabs}

\usepackage{subfigure}

\definecolor{dblue}{RGB}{52, 104, 192}
\definecolor{dgreen}{RGB}{65, 171, 93}
\definecolor{dred}{RGB}{210, 69, 69}
\definecolor{dred2}{RGB}{169, 68, 56}

\def \TAB {Table}
\def \FIG {Figure}

\def \SEC{Section}
\def \APP{Appendix}

\def \FirstOrder{MRF}
\def \EndOrder{LRF}
\def \MiddleOrder{MRC}

\newcommand{\subscript}[2]{$#1 _ #2$}
\usepackage{xcolor,colortbl}
\usepackage{enumitem}
\newcommand{\ie}[0]{\emph{i.e., }}

\newcommand{\eg}[0]{\emph{e.g., }}

\newcommand\sect[1]{\S\ref{#1}}

\title{Understanding the Role of User Profile \\in the Personalization of Large Language Models}

\author{Bin Wu~~~~~~Zhengyan Shi~~~~~~Hossein A. Rahmani~~~~~~Varsha Ramineni\\
        \textbf{Emine Yilmaz}\\
        University College London, United Kingdom \\ 
        \texttt{\{bin.wu.23,emine.yilmaz\}@ucl.ac.uk} \\
}

\begin{document}
\maketitle
\begin{abstract}
Utilizing user profiles to personalize Large Language Models (LLMs) has been shown to enhance the performance on a wide range of tasks. However, the precise role of user profiles and their effect mechanism on LLMs remains unclear. This study first confirms that the effectiveness of user profiles is primarily due to personalization information rather than semantic information. Furthermore, we investigate how user profiles affect the personalization of LLMs. Within the user profile, we reveal that it is the historical personalized response produced or approved by users that plays a pivotal role in personalizing LLMs. This discovery unlocks the potential of LLMs to incorporate a greater number of user profiles within the constraints of limited input length. As for the position of user profiles, we observe that user profiles integrated into different positions of the input context do not contribute equally to personalization. Instead, where the user profile that is closer to the beginning affects more on the personalization of LLMs. Our findings reveal the role of user profiles for the personalization of LLMs, and showcase how incorporating user profiles impacts performance providing insight to leverage user profiles effectively.
\end{abstract}

\section{Introduction}

Large Language Models (LLMs) trained on vast quantities of corpus have greatly advanced the field of natural language processing (NLP) \citep{brown2020language,liu2023chatgpt,chang2023survey}.
Despite their capabilities, LLMs are often equipped with a general level of cognition, which might not fully satisfy the diverse demands of users in practical scenarios \citep{chen2023large, liu2023chatgpt, wu2022meta}. 
This is primarily because preferences can vary significantly among users, leading to situations where the same input can yield different, or even contradictory expectations.
Consequently, there is a clear need for personalizing LLMs to address this issue.

\begin{figure}[!ht]
\centering
\includegraphics[width=\columnwidth]{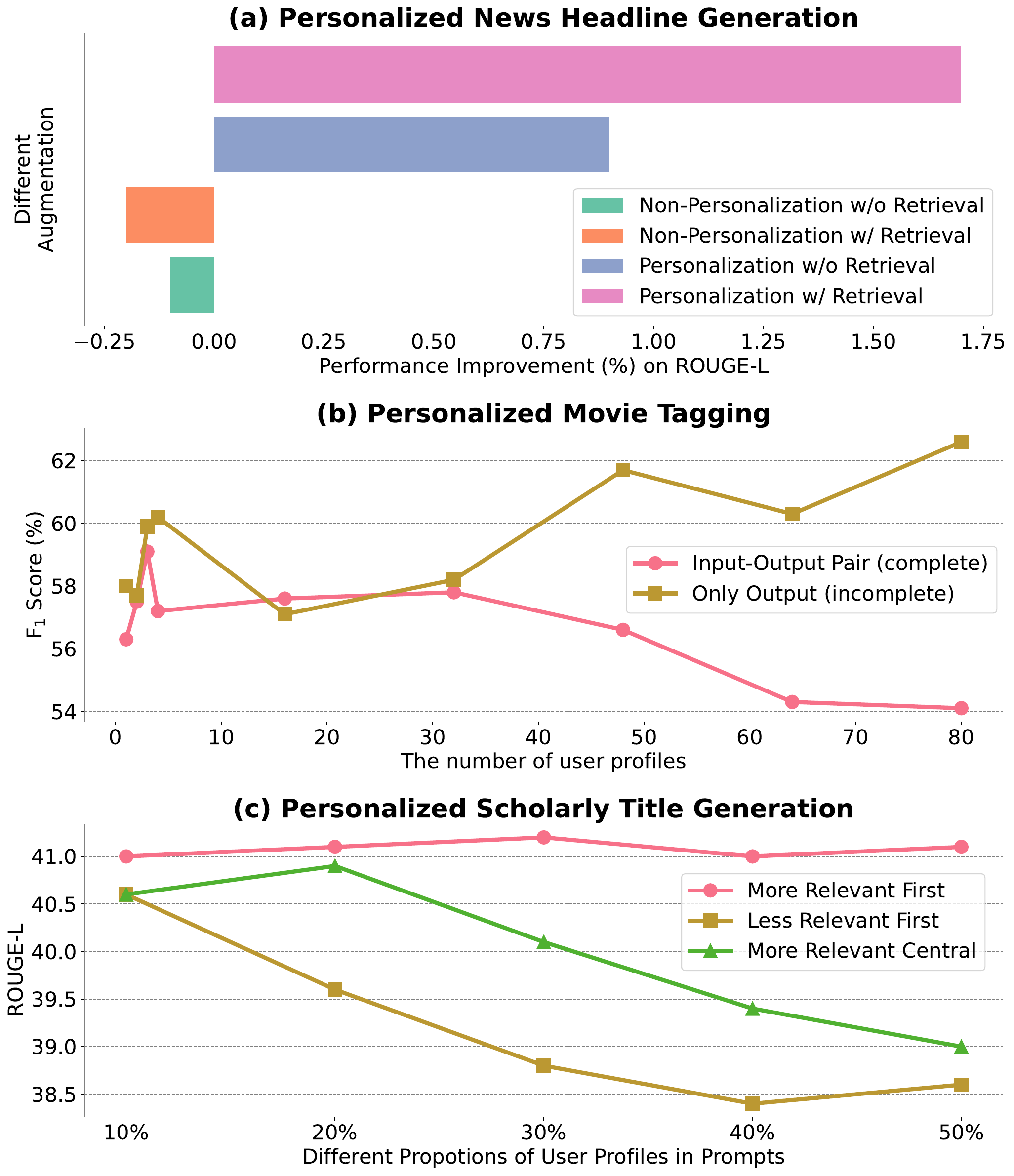}
\caption{Our key findings: \textbf{(a)} Semantic information is less critical to the effectiveness of user profile than personalization, and it only contributes when built on top of personalization (\sect{sec:actual_affect}); \textbf{(b)} The impact of user responses on personalization is greater than that of previous input and their mapping between users' previous input and response (\sect{subsec: single user profiles}); \textbf{(c)} The user profiles in the different positions of the context contribute differently to the personalization, where profiles positioned closer to the beginning contribute more (\sect{subsec: the order}).}
  \label{fig:three_figures}
\end{figure}

One intuitive approach for providing personalized service with LLMs is to utilize user profiles, particularly historical records, to capture underlying user preferences \citep{salemi2023lamp}. By incorporating selected historical records alongside the initial input as context, LLMs can generate personalized responses. Existing research suggests that user profiles can effectively enhance performance across a wide range of tasks by retrieving relevant user profiles from current users \citep{richardson2023integrating, kang2023llms, salemi2024optimization}. However, it remains unclear how incorporated user profiles specifically affect LLM personalization. This uncertainty arises from potential overlaps with retrieval-augmentation generation (RAG) \citep{lewis2020retrieval} and in-context learning (ICL) \citep{brown2020language}, where personalization is not directly addressed. RAG relies on retrieving documents relevant to the input, implicitly requiring semantic information to provide answers \citep{mallen2023not}, while ICL uses input-output pairs as demonstrations to guide LLMs in forming relevant mappings \citep{garg2022can}, necessitating complete input-output pairs. Therefore, the actual role of user profiles in LLM personalization remains ambiguous, complicating the collection and effective use of user information in the real world.

In this paper, we aim to explore the role of user profiles in the personalization of LLMs by answering the following questions: 

\begin{enumerate}[label=\subscript{\textbf{RQ}}{{\arabic*}}]
    \vspace{-0.5em}
    \item \label{Q1} \textit{Do user profiles rely on semantic information (context similar to the input) to improve LLM performance?}
    \vspace{-0.5em}
    \item \label{Q2}  \textit{In what ways do user profiles affect the personalization of LLMs?} 
\end{enumerate}

To address \ref{Q1}, we design and compare the impact of different augmentation methods for user profiles on enhancing performance across various downstream tasks (see \sect{sec:actual_affect}). Our findings indicate that considering only semantic information falls short in improving performance and only contributes to when combined with personalization. Our empirical analysis underscores that personalization information is more critical than semantic information in contributing to user profiles' effectiveness in meeting users' demands.

To address \ref{Q2}, we further investigate how user profiles affect the personalization of LLMs from two perspectives: \textit{within the user profile}, which explores which part of the user profile most significantly contributes to personalizing LLMs (see \sect{subsec: single user profiles}), and \textit{the position of user profiles}, which explores the impact of the order of each user profile in the input context in personalizing LLMs (see \sect{subsec: the order}).
From the perspectives of \textit{within the user profile}, we evaluate the impact of different parts (\eg the previous input or the personalized response), by combining various sampling strategies to construct variants of the user profile.
We find that the personalized response produced or approved by users significantly enhances the personalization, while the previous input and the correct mapping between it and the response are not critical in most cases. 
This finding highlights the potential of LLMs to integrate a great number of user profiles, leading to a significant improvement under the limited input length in our empirical analysis.
From the perspectives of \textit{the position of user profiles} in context,
we surprisingly find that 
(1) user profiles in different positions do not contribute equally to the personalization of LLMs; and that 
(2) the user profiles positioned earlier in the input context tend to have a greater effect on the personalization of LLMs.

In summary, our analysis provides a new view to understand the role of user profiles in personalizing LLMs.
Our findings can be summarized as follows, as shown in \FIG~\ref{fig:three_figures}:
\begin{itemize}
    \vspace{-0.5em}
    \item User profiles rely largely on personalization information rather than semantic information; semantic information contributes to user profiles only when combined with personalization.
    \vspace{-0.5em}
    \item Responses that are generated or approved by users play a crucial role in the personalization of LLMs, while the correct mapping within user profiles is not always essential for effective personalization;
    \vspace{-0.5em}
    \item The impact of user profiles in different positions on personalization is not uniform: the user profile closer to the start of the input context tends to have a larger effect on personalizing LLMs.
\end{itemize}

\section{Related Work}

\paragraph{Personalization of LLMs.}
LLMs have shown impressive performance in different fields \citep{achiam2023gpt, shi2023don}, but there is a gap between models' response and humans' expectations due to general cognition from the pre-training stage. 
Recent efforts, such as reinforcement learning from human feedback (RLHF) \citep{ouyang2022training,yang2024bayesian} to personalize LLMs, try to align LLMs with users' preferences \citep{jang2023personalized, shi2024instruction}, but 
is easily subject to a narrow set of preferences caused by the limited number of crowd-workers \citep{kirk2023Personalisation, wu2023adaptive}. 
Therefore, our work focuses on another line of research, personalizing the LLMs via utilizing user profiles. In this way, \citet{salemi2023lamp} offered a comprehensive evaluation benchmark with diverse language tasks, where the user profiles consist of the user's previous queries and the personalized response. 
More recent works have demonstrated the effectiveness of introduced user profiles in different ways, such as the summarization \citep{richardson2023integrating}, keywords synthesis \citep{li2023teach}, user embeddings \citep{doddapaneni2024user, ning2024user}, parametric knowledge \citep{tan2024democratizing} via Lora \citep{shi2023dept} or the prompt rewriting via reinforcement learning \citep{li2023automatic}. 
While user profiles have been shown to enhance performance on traditional metrics, it is still uncertain which information contributes to the improvement.
The difficulty lies in the overlap with the retrieved documents in RAG, which requires the semantically similar context to provide the direct answer.
Additionally, how introduced user profiles impact the personalization of LLMs is underexplored. 
Therefore, understanding the role of the user profiles is significant and would improve our understanding of the personalization of LLMs. 

\paragraph{Retrieval-Augmentation Generation.}
RAG has become one powerful way to unlock the ability of LLMs to incorporate external knowledge \citep{lewis2020retrieval}. 
The use of retrieved documents mitigates the issue of hallucination \citep{gao2023retrieval}, especially on knowledge-intensive tasks. Various efforts have highlighted the necessity of post-retrieval strategies for effectively utilizing retrieved documents, such as reranking \citep{liu2023lost} and compression \citep{chen2023walking}. 
However, the role of the user profile in personalizing the LLMs is different from the retrieved documents in RAG, where RAG expects semantic information (context similar to input) of the retrieved documents to provide the direct answer, while 
user profiles are designed to reflect user preferences more broadly.

\paragraph{In-Context Learning.}
ICL \citep{brown2020language} takes the input-label pair as the demonstration to prompt the LLMs to deal with downstream tasks. Some efforts are made to explore what factor influences ICL, such as the format of the demonstration \citep{minrethinking, wei2023larger}, and the selection of the demonstration \citep{liu2021makes, gao2023ambiguity}. Some other works aim to explore the inner mechanism of the ICL, including the learning ability \citep{garg2022can, akyurek2022learning}, the Bayesian inference view \citep{xie2021explanation} and the information flow \citep{DBLP:conf/emnlp/WangLDCZMZS23}. 
However, user profiles are different from the demonstrations in ICL, since LLM is expected to capture user preference from the profiles instead of directly formulating the mapping. Therefore, the existing works in ICL cannot directly explain the role of the user profiles in the personalization of LLMs.

\section{Preliminary}

\paragraph{Problem Definition.}
The problem definition of the personalization of LLMs follows LaMP \citep{salemi2023lamp}: given a textual input $x$ describing a user $u$'s task, the goal for the LLM, parameterized by $\boldsymbol{\theta}$, is to generate a personalized response $y$ conditioned on the user $u$. The user $u$ is represented by the user profiles $\mathcal{P}_{u}$ defined as users' historical data that consists of the user input and the personalized response produced or approved by the users, \ie $\mathcal{P}_{u} = \{(x_{u_1}, y_{u_1}),..., (x_{u_{N_u}}, y_{u_{N_u}})\}$ with $N_u$ being the number of the user profiles. Mathmetically, the goal is to formulate the personalized response $p(y|x, \mathcal{P}_{u}; \boldsymbol{\theta})$.
To address it, existing methods mainly include three components: (1) query generation $\phi_q$ to transform the input into the query $q$ for retrieval; (2) retrieval model $\mathcal{R}(q, \mathcal{P}^{\prime}, k)$ to retrieve the relevant $k$ user profile from the candidate $\mathcal{P}^{\prime}$; and (3) prompt construction $\phi_{p}$ to merge the original input $x$ with the retrieved user profiles.

\begin{table}
    \centering
    \resizebox{\columnwidth}{!}{\begin{tabular}{ccc}
        \toprule
         & \bf Task Description & \bf Metrics\\
         \midrule
         LaMP-1& Personalized Citation Identification& Acc\\
         LaMP-2& Personalized Movie Tagging& Acc, $F_1$ Score \\
         LaMP-3& Personalized Product Rating & MAE, RMSE\\
         LaMP-4& Personalized News Headline Generation &ROUGE-1\&-L\\
         LaMP-5& Personalized Scholarly Title Generation&ROUGE-1\&-L\\
         LaMP-7& Personalized Tweet Paraphrasing&ROUGE-1\&-L\\
         \bottomrule
    \end{tabular}}
    \caption{Task description for the used LaMP dataset.}
    \vspace{-1em}
    \label{tab: task description}
\end{table}

\paragraph{Experimental Setup.}
We analyze the personalization of LLMs using the LaMP benchmark \citep{salemi2023lamp} (shown in \TAB~\ref{tab: task description}). Flan-T5-base \citep{chung2022scaling} are employed as the model architecture. More details can be seen in \APP~\sect{app: exp_details}. Our code is available at \href{https://github.com/Bingo-W/Personalisation-in-LLM}{https://github.com/Bingo-W/Personalisation-in-LLM}.

\begin{figure*}[!t]
    \centering
    \includegraphics[width=0.9\textwidth]{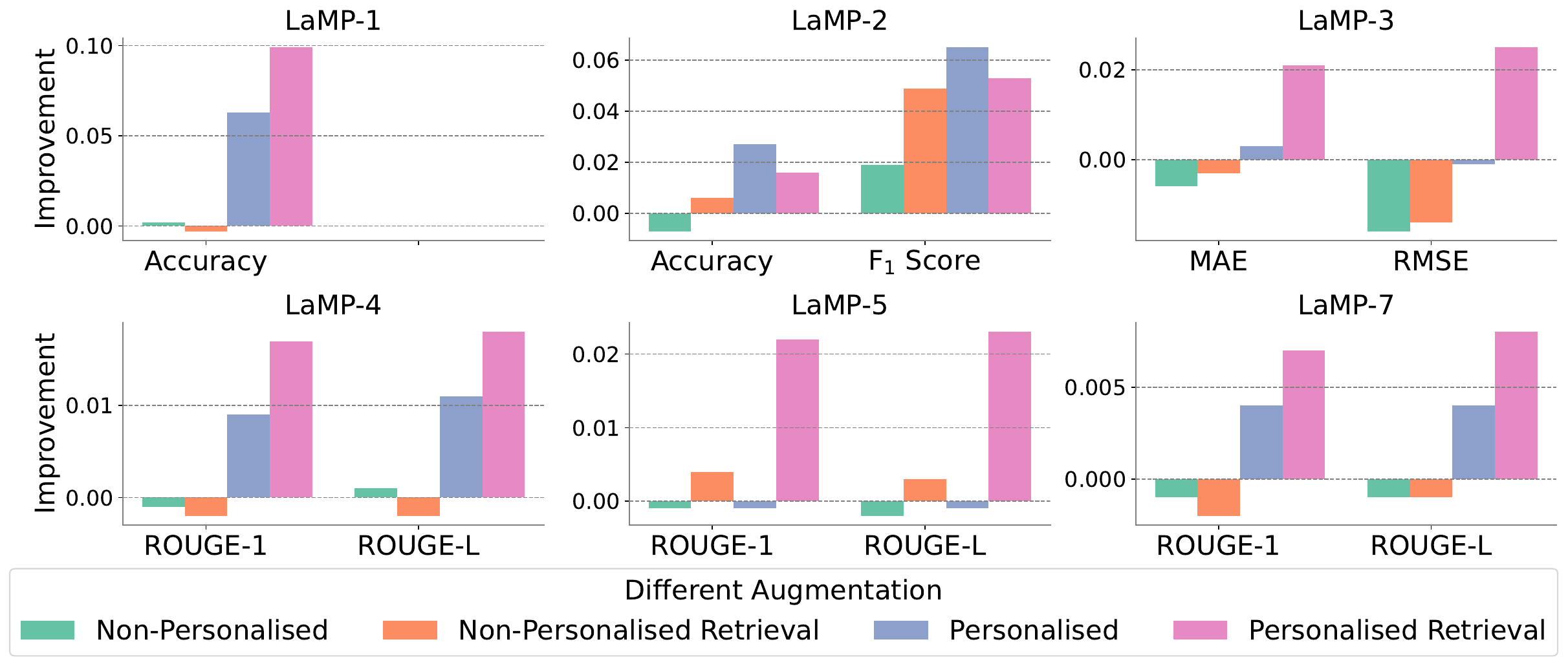}
    \vspace{-0.5em}
    \caption{The improvement of performance (Flan-T5-base) on LaMP dataset with different Augmentation based on the user profiles ($k=1$) compared to without augmentations. 
    Note that LaMP-3 shows a \textbf{decreases} in performance compared to the no-augmentations baseline, indicated by the lower values of both MAE and RMSE, where lower scores signify better performance.}
    \vspace{-1em}
    \label{fig: augmentation for flan-t5}
\end{figure*}

\section{The Actual Affect of User Profile}
\label{sec:actual_affect}
To rigorously assess for user profile, whether semantic information (\ie context similar to the input) or personalization information (context similar to the user) contributes more effectively to the personalization of LLMs (\ref{Q1}), we compare four distinct augmentation strategies employing various methods for sampling user profiles

\noindent\textbf{Non-Personalization Augmentation w/o Retrieval.} 
This strategy involves randomly sampling $k$ profiles from the entire set of user profiles $\mathcal{P}$. This approach does not take into account either personalization or semantic information.

\noindent\textbf{Non-Personalization Retrieval-Augmentation.} In this strategy, a subset of user profiles $\mathcal{P}{sub} \subset \mathcal{P}$ is randomly sampled, with the subset size equal to the average number of profiles per user $\overline{N_u}$, to ensure fair comparison. Retrieval is then employed to select the $k$ most relevant profiles based on the input query $\mathcal{R}(\phi{q}(x_i), \mathcal{P}_{sub}, k)$. This strategy considers only semantic information (\ie retrieving the relevant user profile), ignoring personalization information.

\noindent\textbf{Personalization Augmentation w/o Retrieval.} 
This method randomly samples profiles directly from the current user's set, \ie $\mathcal{P}_{u}$. It takes into account personalization information only (\ie only consider the current user's profile), without considering semantic information.

\noindent\textbf{Personalization Retrieval Augmentation.} 
This strategy retrieves the $k$ most relevant user profiles from the current user's set based on the input query $\mathcal{R}(\phi_{q}(x_i), \mathcal{P}_{u}, k)$. It considers both personalization and semantic information.
It takes into account both personalization and semantic information.

\subsection{Results: Semantic Information Matters Less Than Personalization}
\label{subsec: fine-tune model}
We present the performance improvements with different augmentations compared to scenarios with no user profiles in \FIG~\ref{fig: augmentation for flan-t5}. 

\paragraph{Semantic Information Alone Cannot Lead to Consistent Performance Improvement.}
Considering only semantically similar context (\ie Non-Personalization Retrieval-Augmentation) improves performance on LaMP-2 and LaMP-5 but decreases performance on the remaining four tasks. When compared to considering personalization information only (\ie Personalization Augmentation), the semantic information approach shows a performance gap in most tasks. These observations indicate that semantic information (\ie context similar to the input) has a limited impact on the effectiveness of user profiles in the personalization of LLMs.

\paragraph{Semantic Information contributes to performance improvement Only When Combined with Personalization.}
When the candidate set is limited to user profiles from the current user (\ie context similar to the user), retrieval augmentation (\ie Personalization Retrieval Augmentation) shows performance improvements across almost all tasks. It outperforms strategies that consider either semantic information or personalization alone. This evidence suggests that semantic information contributes to the effectiveness of user profiles only when built on top of personalization information.

\paragraph{Summary.}
Our results validate that semantic information has less impact on the effectiveness of user profiles compared to personalization information, and only contributes when combined with personalization. Additional experiments in \APP~\sect{app: more results about augmentation} confirm that our conclusions hold across different quantities of user profiles and even with larger non-fine-tuned models (\eg \textsc{Llama}-2).

\section{How do User Profile Affect the Personalization of LLMs}
This section further explores how the user profiles influence the personalization of LLMs (\ref{Q2}). 
We specifically explore which components of the user profile contribute to personalizing LLMs, and examine the impact of the position and order of user profiles within the input context on personalization effectiveness.

\subsection{The Effective Part of User Profile}
\label{subsec: single user profiles}
\begin{figure*}[!t]
    \centering
    \includegraphics[width=0.8\textwidth]{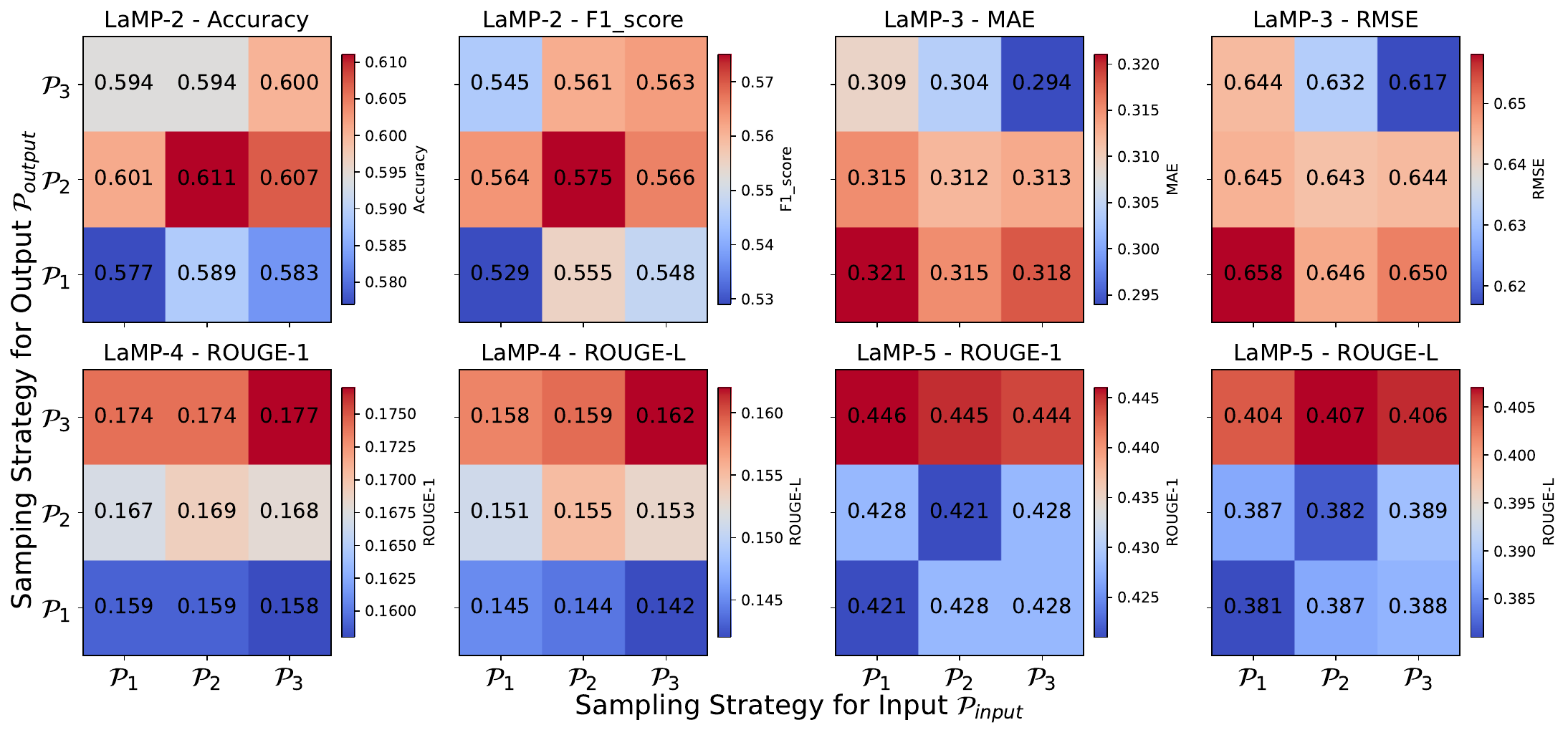}
    \vspace{-0.5em}
    \caption{The performance on 4 LaMP tasks, with different combinations of sampling strategies for the construction of the introduced user profiles. Note that the metrics for LaMP-3 are the lower, the better.}
    \vspace{-1em}
    \label{fig:different construction for user profiles}
\end{figure*}
We identify three key parts within user profiles that potentially affect LLM personalization:
(\texttt{a}) the previous input from users,
(\texttt{b}) the response produced or endorsed by users, and 
(\texttt{c}) the accurate mapping between these inputs and responses.

To investigate the impact of three parts, 
we employ two distinct strategies to sample or retrieve two lists of user profiles, which are then merged to form a final profile list. 
Specifically, user profiles are randomly selected from the entire pool of user profiles, denoted as $\mathcal{P}_{1} \in \mathcal{P}$, or chosen either randomly $\mathcal{P}_{2}$ or semantically based on the retrieval model $\mathcal{P}_{3}$ from the profiles of the current user $\mathcal{P}_{u}$. 
The previous section has demonstrated that the effectiveness of the selected user profile for personalization is ordered by $\mathcal{P}_{3}>\mathcal{P}_{2}>\mathcal{P}_{1}$.
We only use the input part from the first list and the output part from the second list from two strategies, \ie $\mathcal{P}^{\prime} = \{(x_i, y_j)|(x_i, y_i)\in\mathcal{P}_{input}, (x_j, y_j)\in\mathcal{P}_{output}\}$, where $\mathcal{P}_{input}, \mathcal{P}_{output} \in \{\mathcal{P}_{1}, \mathcal{P}_{2}, \mathcal{P}_{3}\}$. 
Note that if the employed sampling strategies for input and output are the same, \ie$\mathcal{P}_{input}=\mathcal{P}_{output}$, it would maintain the correct mapping within the user profiles due to the same list of the selected user profiles. We report the results on four tasks (\ie LaMP-2, LaMP-3, LaMP-4 and LaMP-5) that have the complete user profiles in \FIG~\ref{fig:different construction for user profiles}.

\subsubsection{Results: The Impact of Different Part}

\paragraph{Correct mapping is not necessary for personalization.} 
In \FIG~\ref{fig:different construction for user profiles}, diagonal cells from the bottom-left to the up-right (\ie $\mathcal{P}_{input}=\mathcal{P}_{output}$) maintain the correct mapping, whereas other cells do not. 
However, except for the bottom-left cells, a consistent increase is not evident in centric and up-right cells compared to their respective row (incorrect input) and column (incorrect output) counterparts. 
For instance, in LaMP-5, using personalized w/o retrieval for both input and output performs worse than its row neighbor (0.421 vs.~0.428/0.428) and its column neighbor (0.421 vs.~0.428/0.445).
These observations suggest that the correct mapping between previous input and response is not necessary for LLM personalization. 
An exception is observed in LaMP-3, where the centric and up-right cell performs better, possibly because LaMP-3 uses virtual labels (scores from 1 to 5) without semantic information, thereby necessitating correct mapping to guide LLMs in interpreting the scores accurately.

\paragraph{Previous Input has limited impact on the personalization.} Due to the more powerful sampling strategy from left to right, the right cell uses user profiles with a more effective input for personalization than the left cell.
However, no substantial increase is observed across all four tasks. 
In LaMP-4, with a more powerful input part within user profiles, performance on centric and bottom-centric cells with $\mathcal{P}_{2}$ for the input part is better than their right neighborhood with $\mathcal{P}_{3}$ for the input part (0.155 vs. 0.153 and 0.144 vs. 0.142 for ROUGE-L metric). 
Our results indicate that the improvement does not result from the complete user profile, suggesting that personalization within the previous input has a limited impact on LLM personalization.

\paragraph{The response produced or endorsed by users contributes to the personalization.} 
Cell-related strategies from bottom to top use user profiles with a more effective output for personalization. 
A noticeable change in color from \textcolor{dblue}{\texttt{blue}} to \textcolor{dred2}{\texttt{red}} (\textcolor{dred2}{\texttt{red}} to \textcolor{dblue}{\texttt{blue}} for LaMP-3) indicates that a more effective output markedly enhances performance. 
Even with incorrect mapping, the up-centric cell with a more effective output outperforms the centric cell with correct mapping but a less powerful output. 
Our results underscore that, compared to mapping and the input part of user profiles, personalization within the response produced or endorsed by users plays a crucial role in LLM personalization.

\paragraph{Summary.}
It is the response produced or endorsed by users that enhances LLM personalization, rather than correct mapping or previous input from users. Correct mapping is deemed necessary only for non-semantic output space.

\subsubsection{Results: The Impact of Only Output}

Previous findings indicate that the correct mapping between input and output does not necessarily contribute to LLM personalization. 
To further investigate the role of its input and output parts, we conduct experiments where we focus solely on either the input or output part while ignoring the mapping. 
We modify the template part of the prompt with minimal changes (see \APP~\sect{app: prompt}). 
It is important to note that using the incomplete user profile helps to shorten the input context, thereby enabling the incorporation of more user profiles for personalizing LLMs within the limited input length. 
We expand the number of utilized user profiles, ranging from $10\%$ to $50\%$ of the most relevant profiles. 

\paragraph{Only using the output part substantially enhances the personalization.}
\begin{figure*}[!t]
    \centering
    \vspace{-1em}
    \includegraphics[width=0.8\textwidth]{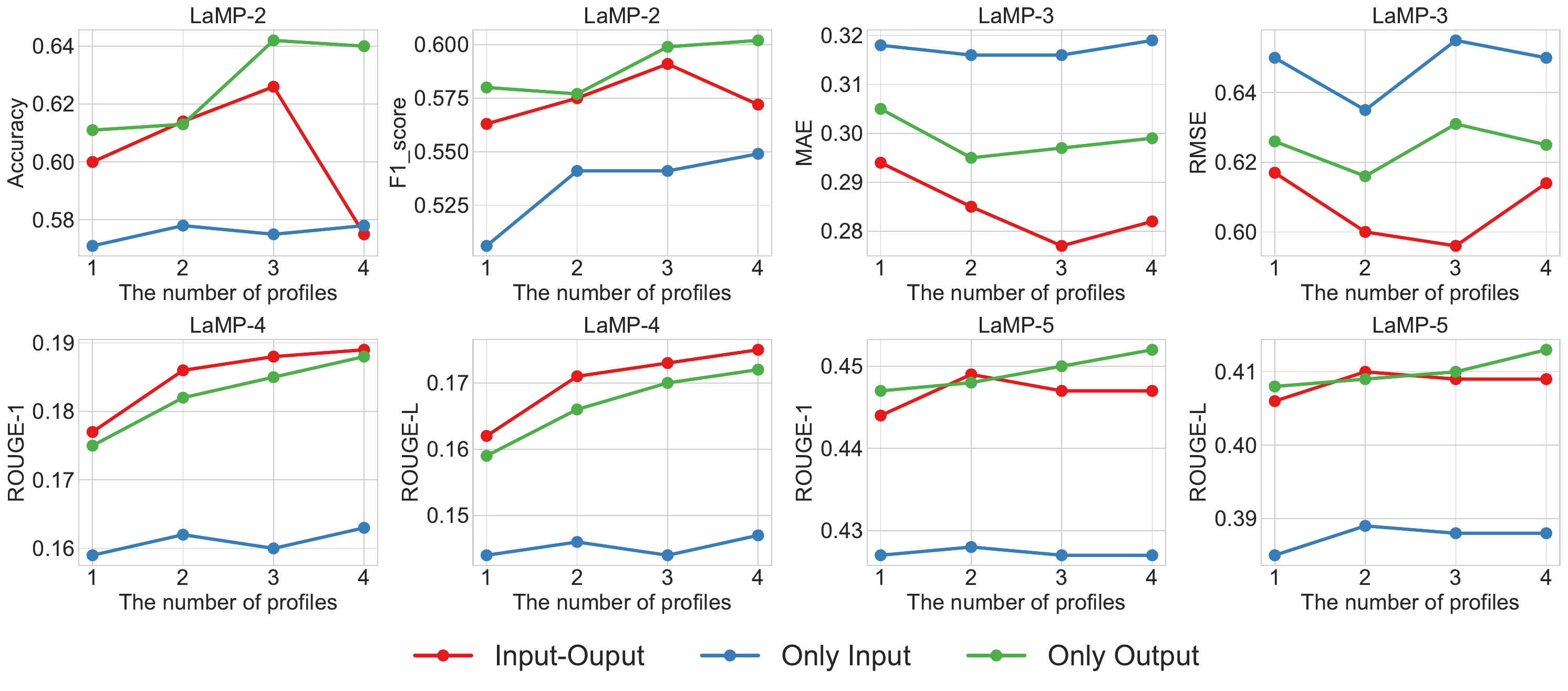}\
    \vspace{-1em}
    \caption{The performance with different numbers of used user profiles on four LaMP datasets when only with the input part and only with the output part. Note that the metrics for LaMP-3 are the lower, the better.}
    \vspace{-0.5em}
    \label{fig: only input and output}
\end{figure*}
Our results in \FIG~\ref{fig: only input and output} reveal that, except for LaMP-3, using only the output part of user profiles achieves comparable (LaMP-4) or even superior performance (LaMP-2 and LaMP-5) compared to using complete user profiles. 
In contrast, utilizing only the input part leads to noticeable performance degradation across all tasks. 
This supports our earlier findings, emphasizing the importance of user profiles with a more powerful output for LLM personalization. 
It further underscores that responses produced or endorsed by users play a pivotal role in effective personalization, particularly when contrasted with correct mapping and previous input considerations. 
In the case of LaMP-3, where tasks involve non-semantic label spaces, utilizing incomplete user profiles (only output or input) leads to a noticeable decrease, supporting the notion that correct mapping is essential for such tasks.

\paragraph{Only using output unlocks the potential to use more user profiles.}

\begin{figure*}
    \centering
    \includegraphics[width=0.8\textwidth]{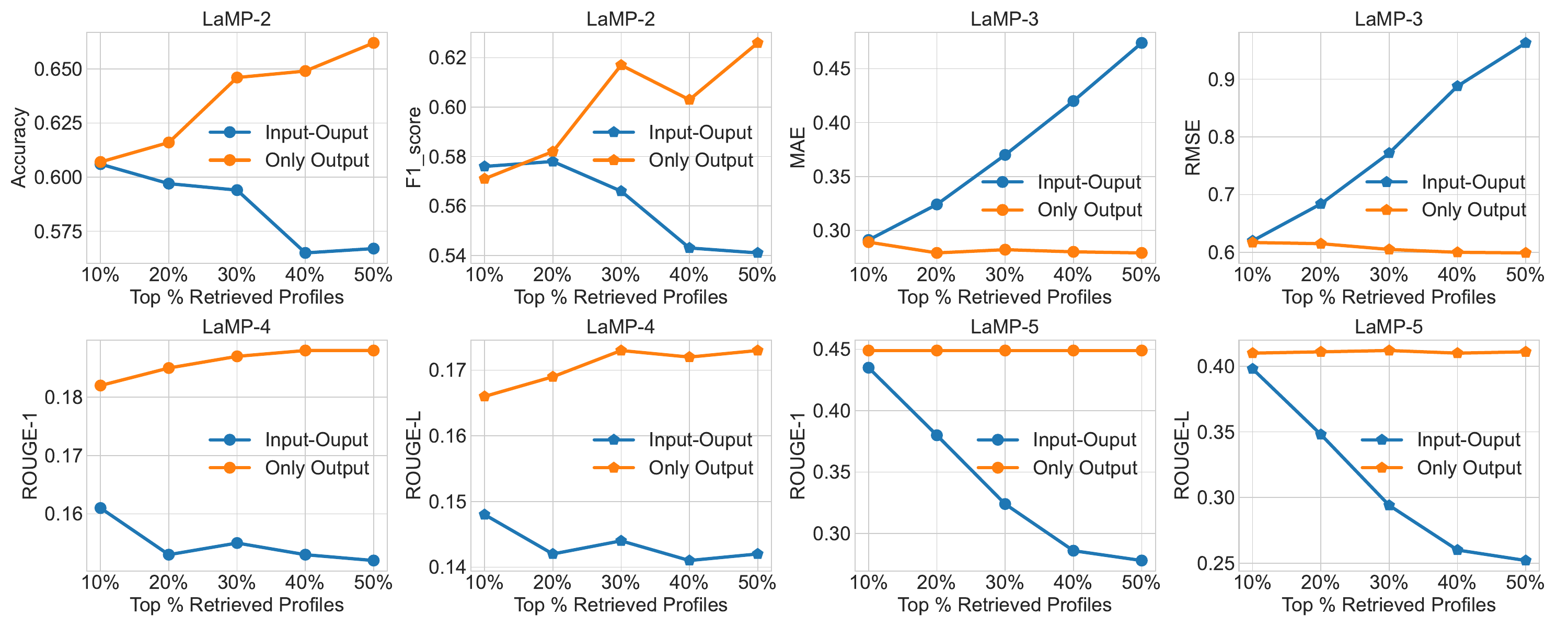}
    \vspace{-0.5em}
    \caption{The performance with different proportions of user profiles on four LaMP datasets when only using the output part of the completed user profile. Note that the metrics for LaMP-3 are the lower, the better.}
    \vspace{-1em}
    \label{fig: only input and output with different proportion.}
\end{figure*}

As depicted in \FIG~\ref{fig: only input and output with different proportion.},
our analysis of various proportions of user profiles reveals that using complete user profiles often exceeds the maximum input length, necessitating truncation that leads to substantial performance losses. In contrast, focusing solely on personalized responses improves performance across all tasks compared to using complete user profiles. Notably, there is a substantial increase in LaMP-2 (from 0.571 to 0.626 for the F$_1$ score) and LaMP-4 (from 0.182 to 0.188 for ROUGE-1). 
For LaMP-3 and LaMP-5, where using complete user profiles suffers from severe degradation, exclusively leveraging the output of user profiles proves to be a robust strategy. This not only affirms the efficacy of the output part for personalization but also highlights that using only the output part extends the capacity of LLMs to leverage more user profiles, leading to greater performance improvement.

\paragraph{Summary.}
The findings underscore the vital contribution of the output part of user profiles to LLM personalization. 
This finding supports the strategy of extending the LLMs' capacity to utilize more user profiles, especially when focusing on the output, which may result in notable performance gains.

\subsection{Exploring the Impact of User Profile Order}
\label{subsec: the order}
The order in which multiple user profiles are arranged within the input context may largely affect the personalization capabilities of LLMs. 
Previous work in RAG \citep{liu2023lost} has indicated that LLMs may overlook the document containing the correct answer when placed in the middle position of the input context. 
Considering that user profiles differ fundamentally from the documents typically retrieved in traditional RAG setups, it's crucial to assess how the order of user profiles influences personalization.

To explore the effect of varying orders of user profiles in the input context, we select the top $k$ relevant user profiles and arrange them in the input context using different approaches (seen \TAB~\ref{tab: example for different order}):

\noindent\textbf{More Relevant First (\FirstOrder{}).} 
Following conventional retrieval practices \citep{salemi2023lamp}, this method places profiles with higher semantic similarity earlier in the input context.

\noindent\textbf{Less Relevant First (\EndOrder{}).} 
This order is the inverse of the More Relevant First, positioning the most relevant profiles nearer to the end of the context.

\noindent\textbf{More Relevant Central (\MiddleOrder{}).} Unlike the above orders and the ListInMiddle Ranker in RAG \citep{liu2023lost}, this method places the more relevant user profiles in the more central position.

\label{subsec: multiple profiles}
\begin{figure*}
    \centering
    \vspace{-1em}
    \includegraphics[width=\textwidth]{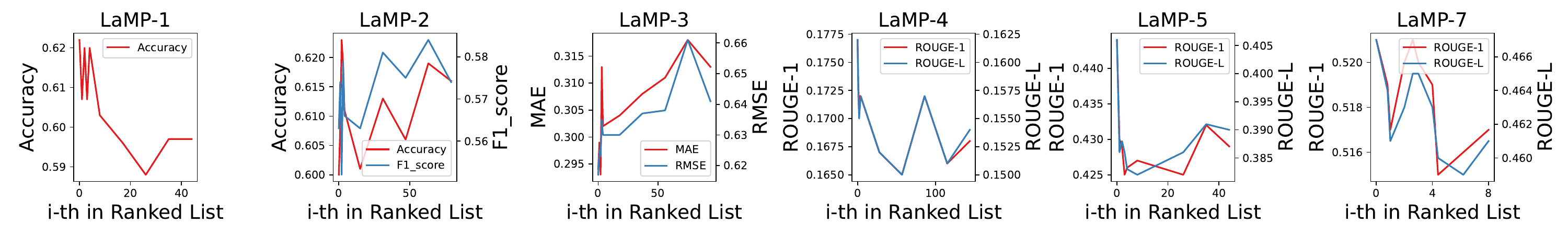}
    \vspace{-2em}
    \caption{The performance on LaMP with user profiles located in different positions in the retrieved ranked list. Note that $.1$ refers to the $10\%$-th position. Note that the metrics for LaMP-3 are the lower, the better.}
    \vspace{-0.5em}
    \label{fig: top i-th in the retrieval}
\end{figure*}

\begin{table}[!t]
    \centering
    \tiny
    \resizebox{0.9\columnwidth}{!}{\begin{tabular}{cc}
        \toprule
        \bf Ordering Strategy & \bf User Profile Order \\
        \midrule
        BM25 Original Order & [1], [2], [3], [4], [5]\\
        More Relevant First & [1], [2], [3], [4], [5]\\
        Less Relevant First & [5], [4], [3], [2], [1]\\
        More Relevant Central & [4], [2], [1], [3], [5] \\
        \bottomrule
    \end{tabular}}
    \vspace{-0.5em}
    \caption{Illustration of the order of user profiles in our experiments, where [1] refers to the most relevant one according to the BM25 relevance ranking.}
    \vspace{-2em}
    \label{tab: example for different order}
\end{table}

\begin{table}[!t]
    \centering
    \resizebox{0.9\columnwidth}{!}{\begin{tabular}{ccccc}
        \toprule
         && \bf \EndOrder{} & \bf \MiddleOrder{} & \bf \FirstOrder{} \\
         \midrule
         LaMP-1&Acc.$\uparrow$&0.707&\textbf{0.714}&0.711  \\
         \hline
         \multirow{2}{*}{LaMP-2}&Acc.$\uparrow$&0.606&\textbf{0.611}&0.606  \\
         &F1$\uparrow$&0.568&\textbf{0.584}&0.576 \\
         \hline
         \multirow{2}{*}{LaMP-3}&MAE $\downarrow$&0.296&0.298&\textbf{0.291} \\
         &RMSE $\downarrow$&0.628&0.632&\textbf{0.620}\\
         \hline
         \multirow{2}{*}{LaMP-4}&ROUGH-1$\uparrow$&0.152&0.154&\textbf{0.161}\\
         &ROUGH-L$\uparrow$&0.140&0.142&\textbf{0.148}\\
         \hline
         \multirow{2}{*}{LaMP-5}&ROUGH-1$\uparrow$&0.428&0.430&\textbf{0.435}\\
         &ROUGH-L$\uparrow$&0.391&0.393&\textbf{0.398}\\
         \hline
         \multirow{2}{*}{LaMP-7}&ROUGH-1$\uparrow$&0.520&0.520&\textbf{0.524}\\
         &ROUGH-L$\uparrow$&0.465&0.466&\textbf{0.469}\\
         \bottomrule
    \end{tabular}}
    \vspace{-0.5em}
    \caption{The performance with different ordering strategies for the top $10\%$ user profiles on LaMP datasets. The best result for each dataset is highlighted in bold.}
    \vspace{-1em}
    \label{tab: result with different orders}
\end{table}

\begin{figure*}[!t]
    \centering
    \includegraphics[width=0.88\textwidth]{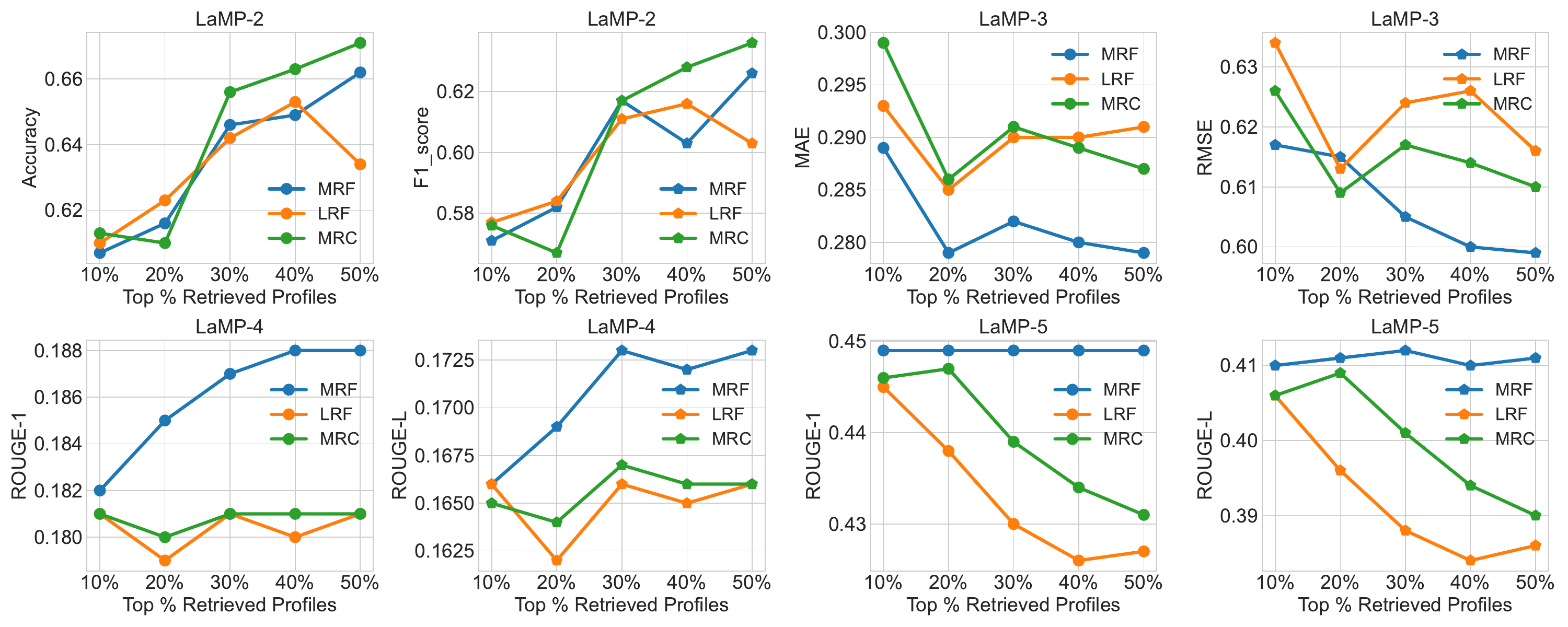}
    \vspace{-0.5em}
    \caption{The performance with different proportions of user profiles on four LaMP datasets when only using the output part with three different orders. Note that the metrics for LaMP-3 are the lower, the better.}
    \vspace{-1em}
    \label{fig: only output with order}
\end{figure*}

\subsubsection{Results: The Impact of Different Order}

\paragraph{User profiles in different positions contribute not equally to the personalization.}
Our results in \TAB~\ref{tab: result with different orders} reveal substantial variations in performance across tasks when the same subset of user profiles is arranged in different orders within the input context. 
For instance, on LaMP-2, \MiddleOrder{} achieves a MAE of 0.298, whereas \FirstOrder{} records 0.291. Similarly, on LaMP-5, \EndOrder{} yields a ROUGH-1 score of 0.152, while \FirstOrder{} achieves 0.161. This observation emphasizes that the user profiles in different positions of the input context do not contribute equally to performance gains.

\paragraph{The user profile closer to the start of the input context tends to have a larger effect.}
Before we analyze the difference in the position of input context, 
the findings in Figure \ref{fig: top i-th in the retrieval} highlight the importance of semantic relevance, showing that user profiles with higher semantic similarity have a stronger impact on personalization. 
\TAB~\ref{tab: result with different orders} further supports that \FirstOrder{} achieves the best result on all tasks except for LaMP-1 and LaMP-2. \FirstOrder{} exhibits improvement ranging from $1\%$ to nearly $5\%$ across the last four tasks, with a slight decrease on LaMP-2 due to the effectiveness of most user profiles within the current user. Additionally, both \FirstOrder{} and \MiddleOrder{} consistently outperform \EndOrder{} on all tasks, emphasizing that user profiles closer to the beginning contribute more to personalization.

\paragraph{More forward more contribution when only using output.} 
We further examine if the conclusion holds when only using the output part. 
We analyze performances with three different orders only using the output part of different proportions of the top relevant user profiles, ranging from $10\%$ to $50\%$ in \FIG~\ref{fig: only output with order}. 
Our results demonstrate that \FirstOrder{} outperforms on all tasks except LaMP-2 when only using the output part, and the performance gap widens as more user profiles are utilized. 
For LaMP-2, \MiddleOrder{} performs better than \FirstOrder{},
suggesting that profiles placed at the start position (even if they are not the most relevant) substantially contribute to personalization.
This underscores that even when only using the output part of user profiles, the user profile closer to the start of the input context contributes more to the personalization of LLMs.

\vspace{-0.5em}
\section{Conclusion}

In this work, we study the role of user profiles on the LLM personalization.  Our findings indicate that the semantic information contained in user profiles does not significantly contribute to LLM personalization and only has an impact when built on top of personalization. We also reveal that precise input-response mapping is often unnecessary for effective personalization; instead, responses that users produce or endorse are crucial. Moreover, using only the response part of profiles not only matches but can exceed the performance of full profiles when considering input constraints. Additionally, we find that the position of user profiles within the input context largely affects LLM focus, with profiles closer to the start having a greater impact. This study reveals how the user profile affects the personalization of LLMs not only underscores the importance of personalization in LLMs but also clarifies the roles of user profiles in effective personalization, providing insights into optimal utilization strategies.

\clearpage
\section*{Limitations}
While our study provides insights into LLM personalization, it has two key limitations. Firstly, while our research scope has explored various tasks including text classification tasks and sentence-level text generation, we leave the exploration of broader recommendation tasks and paragraph-level text generation for future work. Secondly, due to hardware constraints, our analysis was limited to models with less than 1 billion parameters for fine-tuned settings and 7 billion parameters for frozen models. Future research employing larger-scale models and a wider range of tasks could refine our conclusions and broaden the applicability of our findings.

\section*{Ethics Statement}
The datasets and models employed in this study are sourced from publicly available and open-source repositories. Notably, the dataset provider has already taken measures to anonymize personally identifiable information, mitigating potential ethical concerns. As a result, we do not anticipate any ethical issues arising from the utilization of these datasets. Our commitment to ethical research practices remains paramount, and we ensure compliance with relevant guidelines and regulations throughout the study.

\bibliography{main}

\clearpage
\appendix

\section*{Appendix Overview}
\label{sec:appendix}
The appendix is structured as follows:
\paragraph{Appendix \sect{app: dataset}} provides a brief description for each dataset.
\paragraph{Appendix \sect{app: exp_details}} provides implementation details and hyperparameters used in our experiments.
\paragraph{Appendix \sect{app: prompt}} describes the prompts used in our experiments.
\paragraph{Appendix \sect{app: more results about augmentation}} provides additional experiments regarding different augmentations.

\section{The Details of Dataset}
\label{app: dataset}

We give a detailed introduction to the dataset we used in the following. LaMP \citep{salemi2023lamp} consists of six tasks, from the text classification to the text generation tasks. In our work, we use the user-based dataset and choose six LaMP tasks except LaMP-6 considering the public availability. The details of the selected tasks are:
\begin{itemize}
    \item \textbf{LaMP-1: Personalized Citation Identification} Given a paper from a user, an LLM needs to predict which one of the two candidates will cite in this paper based on the user profile. User profiles refer to the paper that the user has authorized before.
    \item \textbf{LaMP-2: Personalized Movie Tagging} Given a movie description, an LLM needs to predict which one of the 15 candidate tags the user will give to the movie based on the user profile. User Profiles refer to the user’s historical tagging behavior, consisting of the movie description and the given tag.
    \item \textbf{LaMP-3: Personalized Product Rating} Given a review from a user, an LLM needs to predict the score with an integer with the range from 1 to 5 that the user will give based on the user profile. User profiles refer to the user's historical rating behavior, consisting of the reviews and the associated rating score.
    \item \textbf{LaMP-4: Personalized News Headline Generation} Given an article from a user, an LLM needs to generate the headline for this article based on the user profile. User profiles refer to the authors’ historical article-title pairs.
    \item \textbf{LaMP-5: Personalized Scholarly Title Generation} Given an article's abstract, an LLms needs to generate the scholar title for this article based on the user profile. User profiles refer to the user's historical article-title pairs.
    \item \textbf{LaMP-7: Personalized Tweet Paraphrasing} Given a tweet, an LLM needs to generate a tweet in the style of the user. User profiles refer to the user's historical tweets.
\end{itemize}

\paragraph{Evaluation} Following the previous work \citep{salemi2023lamp}, we employ Accuracy for LaMP-1, Accuracy and F1 score for LaMP-2, MAE and RMSE for LaMP-3 and Rough-1 and Rough-L \citep{lin2004rouge} for the left three tasks (\ie LaMP-4, LaMP-5 and LaMP-7).

\paragraph{Statistic Information} We use the average number of user profiles per user in our analysis, which is different across different tasks. The average number of user profiles per user is $90.61$, $159.29$, $188.10$, $287.16$, $89.61$ and $17.74$ for the used six tasks, respectively.

\section{The Details of Experiments}
\label{app: exp_details}
We analyze the personalization of LLMs using the LaMP benchmark \citep{salemi2023lamp}, except LaMP-6 due to unavailability. Performance is reported on the validation dataset as the test dataset is not available. The task description and metrics are shown in \TAB~\ref{tab: task description} and more details are in \APP~\sect{app: dataset}. Following the existing work \citep{salemi2023lamp}, the fine-tuning setting employs Flan-T5-base \citep{chung2022scaling} with 250M parameters, while the non-fine-tuning setting uses the chat version of the Llama 2-7B model \citep{touvron2023llama} for reproducibility. The fine-tuning setting, following the same optimization from \citet{salemi2023lamp}, employs the AdamW optimizer, learning rate of $5\times10^{-5}$, $5\%$ warmup steps with a linear scheduler, and decay of $10^{-4}$.
For text classification tasks (LaMP-1, LaMP-2, LaMP-3), the model is trained for 10 epochs; for text-generation tasks (LaMP-4, LaMP-5, LaMP-7), it's trained for 20 epochs. 
The maximum length for input is set as 512 and for output is set as 128. For the frozen large-scale model, we set the maximum input length as 1500 and the output length as 128. 
Maximum input/output lengths are set at 512/128 for the fine-tuned Flan-T5-base model and 1500/128 for the frozen Llama 2-7B-chat model.
Both settings use a beam search of size 4. All experiments run on a Titan GPU with 24GB memory.

\begin{table*}[!t]
    \centering
    \resizebox{\textwidth}{!}{\begin{tabular}{cccc}
        \toprule
         Tasks& Strategy& Template Part of Prompt & Original Template Part of Prompt \\
         \midrule
         \multirow{2}{*}{LaMP-2}& Only Input& \textit{"The previous movies are \textcolor{red}{[Input]}"}&\multirow{2}{*}{\textit{"The tag for the movie: \textcolor{red}{[Input]} is \textcolor{red}{[Output]}"}}\\
         & Only Output& \textit{"The tag for the previous movies are \textcolor{red}{[Output]}"}&\\
         
         \multirow{2}{*}{LaMP-3}& Only Input& \textit{"The previous reviews are \textcolor{red}{[Input]}"}&\multirow{2}{*}{\textit{"\textcolor{red}{[Output]} is the score for \textcolor{red}{[Input]}"}}\\
         & Only Output& \textit{"\textcolor{red}{[Output]} are the score for the previous review"}&\\
         
         \multirow{2}{*}{LaMP-4}& Only Input& \textit{"The previous articles are \textcolor{red}{[Input]}"}&\multirow{2}{*}{\textit{"\textcolor{red}{[Output]} is the title for \textcolor{red}{[Input]}"}}\\
         & Only Output& \textit{"\textcolor{red}{[Output]} are the title for previous articles"}&\\
         
         \multirow{2}{*}{LaMP-5}& Only Input& \textit{"The previous papers are \textcolor{red}{[Input]}"}&\multirow{2}{*}{\textit{"\textcolor{red}{[Output]} is the title for \textcolor{red}{[Input]}"}}\\
         & Only Output& \textit{"\textcolor{red}{[Output]} are the title for the previous papers"}&\\
         \bottomrule
    \end{tabular}}
    \caption{The template part of the used prompt for only using the input or output part of the user profiles. \textcolor{red}{[Input]} and \textcolor{red}{[Output]} refer to the input part and output part of the used user profiles, respectively.}
    \label{tab: the used prompt}
\end{table*}
\begin{figure*}[!t]
    \centering
    \includegraphics[width=0.9\textwidth]{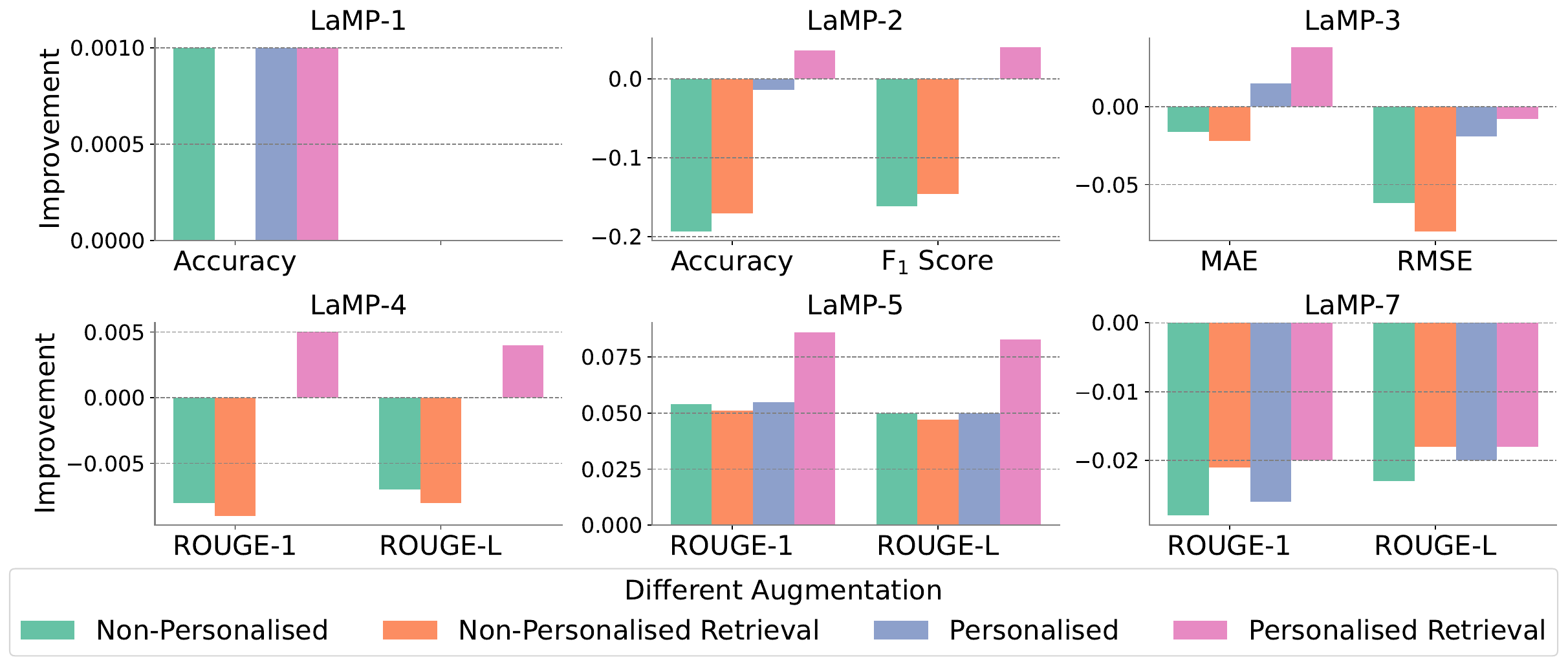}
    \caption{The improvement of performance (Llama 2) on LaMP dataset with different Augmentation based on the user profiles ($k=1$) compared to without augmentations. Note that LaMP-3 shows the \textbf{decreases} when compared to without augmentations since both MAE and RMSE are the lower, the better.}
    \label{fig: augmentation for Llama 2}
\end{figure*}

\begin{figure*}[!t]
    \centering
    \includegraphics[width=\textwidth]{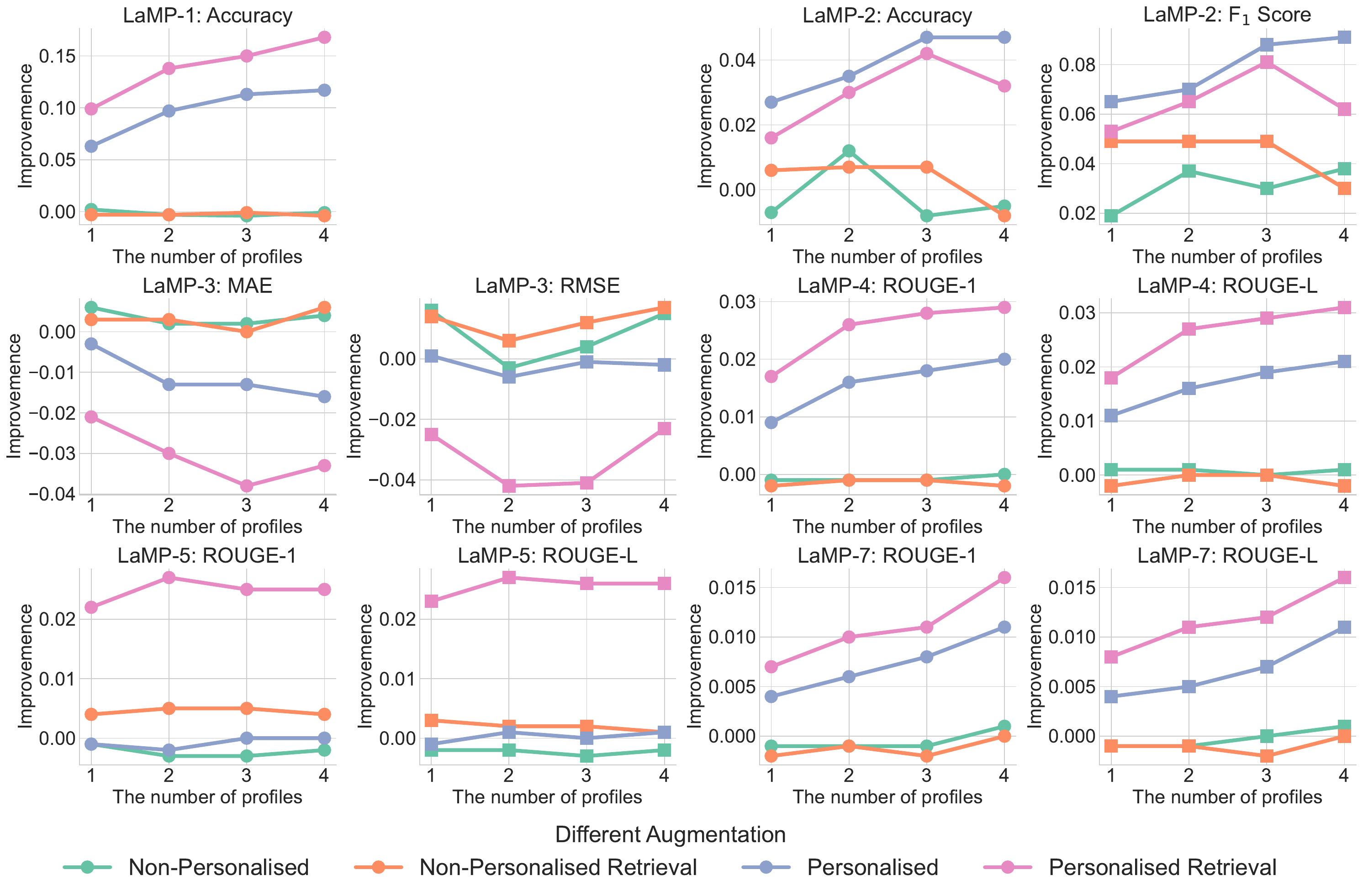}
    \caption{The improvement of performance (Flan-T5-base) on LaMP dataset with Augmentation based on the different number of user profiles compared to without augmentations. Note that LaMP-3 shows the \textbf{decreases} when compared to without augmentations since both MAE and RMSE are the lower, the better.}
    \label{fig: augmentation with different number of used user profiles for flan-t5}
\end{figure*}

\section{The Details of Used Prompts}
\label{app: prompt}

There are two types of prompts that we used in our work. The first one following the previous work \citep{salemi2023lamp} is to utilize the complete user profile. When only input or outpart of the user profile in LaMP-2, LaMP-3, LaMP-4 and LaMP-5, we instead use other prompts with the minimum change compared to the original one. The details can be seen in \TAB~\ref{tab: the used prompt}.


\section{More results about Different Augmentations}
\label{app: more results about augmentation}

\subsection{Results: More User Profile.}
We also show the performance of the augmentation methods with different numbers $k$ of the used user profiles in \FIG~\ref{fig: augmentation with different number of used user profiles for flan-t5}.

\paragraph{The conclusion that user profiles actually enhance personalization remains when with different numbers of user profiles.} The reported results show that compared to non-personalization augmentation methods, the personalization augmentation methods achieve a constant improvement on most tasks, even with the different number of used user profiles. On the other hand, the non-personalization augmentation cannot introduce a constant benefit for the performance. It further confirms the conclusion we get in the main paper.

\paragraph{The enhancement of personalization augmentation increases with more user profiles.} The extent of the enhancement on personalization by personalization augmentation increases from $k=1$ to $k=4$. Especially for the personalization with retrieval sees the improvement on all tasks. However, when more user profiles are utilized, it easily introduces some noise and suffers from degradation. This observation further confirms that within the reasonable range, the enhancement extent by user profile increases with the increased number of used user profiles.

\subsection{Results: Frozen Models}
We also report the results with a frozen Llama 2 with the chat version in \FIG~\ref{fig: augmentation for Llama 2}, to examine the conclusion from \SEC~\ref{subsec: fine-tune model}.
\paragraph{Even for frozen LLMs, user profiles actually enhance personalization, but the performance improvement decreases due to no fine-tuning.}
Results on frozen LLMs exhibit similar patterns, where both two personalized augmentation methods can achieve an improvement on most tasks (except LaMP-7). Conversely, the augmentations without considering the personalization decreases the performance on most tasks, although they positively impact only on LaMP-1 and LaMP-5. 
These observations underscore the pivotal role of user profiles containing user preferences, supporting the assumption that introduced user profiles enhance LLM personalization for performance improvement.
However, it's noteworthy that personalization augmentation with non-fine-tuning improves performance less than the ones with fine-tuning and even introduces some noises on LaMP-7. This suggests that although user profiles can enhance personalization, LLMs without fine-tuning struggle to capture user preferences from the introduced user profile in the input context to obtain performance improvements.

\end{document}